\documentclass[pdflatex,sn-mathphys-num]{sn-jnl}

\usepackage{graphicx}%
\usepackage{multirow}%
\usepackage{amsmath,amssymb,amsfonts}%
\usepackage{amsthm}%
\usepackage{mathrsfs}%
\usepackage[title]{appendix}%
\usepackage{xcolor}%
\usepackage{textcomp}%
\usepackage{manyfoot}%
\usepackage{booktabs}%
\usepackage{algorithm}%
\usepackage{algorithmicx}%
\usepackage{algpseudocode}%
\usepackage{listings}%
\usepackage{microtype}

\usepackage{glossaries}
\newacronym{zmp}{ZMP}{Zero Moment Point} 
\newacronym{mpc}{MPC}{Model Predictive Control} 
\newacronym{srbd}{SRBD}{Single Rigid Body Dynamics}
\newacronym{grf}{GRF}{ground reaction force} 
\newacronym{com}{CoM}{center of mass} 
\newacronym{ud}{UD}{underdamped} 
\newacronym{cd}{CD}{critically damped} 
\newacronym{od}{OD}{overdamped} 
\newacronym{fft}{FFT}{Fast Fourier Transform}
\newacronym{rms}{RMS}{root-mean-square}
\newacronym{armpc}{ARMPC}{arm-aware MPC}
\newacronym{ocp}{OCP}{Optimal Control Problem}
\newacronym{lf}{LF}{Left Front}
\newacronym{lh}{LH}{Left Hind}
\newacronym{rf}{RF}{Right Front}
\newacronym{cdc}{CDC}{cross-diagonal crawl}
\newacronym{bdc}{BDC}{back-diagonal crawl}
\newacronym{cc}{CC}{circular crawl}

\theoremstyle{thmstyleone}%
\theoremstyle{thmstyletwo}%

\theoremstyle{thmstylethree}%

\begin{document}

\title[Article Title]{Gait-Dependent Effects on Quadruped Locomotion for Load-Carrying using Passive Mechanisms}


\author*{\fnm{Giovanni B.} \sur{Dessy}}

\author{\fnm{Claudio} \sur{Semini}}

\author{\fnm{Victor} \sur{Barasuol}}

\affil{%
\small
\parbox{\textwidth}{\centering
Dynamic Legged Systems, Istituto Italiano di Tecnologia, Genoa/GE, 16163, Italy,\\[-1pt]
\texttt{name.surname@iit.it,}\\[-1pt]
\texttt{dls.iit.it}
}%
}



\abstract{Passive mechanical interfaces offer a lightweight alternative to actuated manipulators for quadruped payload carrying, but their impedance directly couples the payload dynamics with the locomotion pattern. This paper analyzes how passive-arm stiffness-damping selection affects payload-carrying locomotion under different gait and payload conditions. We compare damped and underdamped passive-arm impedance configurations in simulation during flat-ground locomotion. For crawl gaits, where the support polygon remains well defined, the results show that underdamped impedance increases passive-joint oscillations and can reduce the ZMP margin with respect to the support polygon. Trot is retained as a dynamic excitation case for the passive arm, but it is not used for direct ZMP-margin stability comparison. The results are summarized in gait-payload-stiffness-damping maps, where ZMP-margin reduction is evaluated for crawl gaits and trot is retained only as a passive-arm excitation case.}




\maketitle

\section{Introduction and Related Work}\label{sec1}
Physical interaction with payloads or human partners is an important capability for legged robots operating in unstructured environments. In addition to navigation and inspection, quadrupeds are increasingly being used to transport  objects \cite{turrisi2026binwalkerdevelopmentfieldevaluation}, collaborate with other agents \cite{pandit2026multiquadrupedcooperativeobjecttransport}, and support manipulation or carrying tasks in environments where wheeled platforms are limited \cite{CorosWheeledPlatforms, carryinguncarriableArashLab}. These applications require the robot to maintain locomotion stability while exchanging forces with a carried load or with a collaborating partner. 

A common way to enable such interaction is to equip quadruped robots with active manipulators and to control the resulting system through whole-body or model predictive loco-manipulation frameworks~\cite{RisiglioneImpedanceControlQuadruped2022,vincentiCentralizedModelPredictive2023,anCollaborativeLocoManipulationPickandPlace2025}. These approaches provide high versatility and allow the robot to regulate interaction forces actively. However, actuated manipulators increase mechanical complexity, mass, power consumption, control requirements and cost. For carrying tasks where dexterous manipulation is not the primary objective, simpler physical interfaces can be preferable. 

Several works have therefore addressed payload transportation using non-dexterous or non-fully-actuated interfaces.
Examples include cable-towed payloads~\cite{yangCollaborativeNavigationManipulation2022}, rigid or semi-rigid connections between multiple quadrupeds \cite{CentrvsDecKim}, and human-robot co-carrying interfaces \cite{suctioncup}. These methods reduce the need for a fully actuated arm, but they also introduce an interaction coupling between the robot and the payload motion. As a result, the load cannot always be treated as a purely static mass attached to the robot. 
The influence of physical interaction on locomotion has also been studied in humanoid collaborative carrying. Agravante et al. proposed walking pattern generators that account for sustained interaction forces during human-humanoid carrying tasks~\cite{agravanteWalkingPatternGenerators2016}, and later integrated these ideas into a complete human-humanoid collaborative carrying framework~\cite{agravanteHumanHumanoidCollaborativeCarrying2019a}. These works show that interaction forces can be considered directly in locomotion generation. In contrast, the present paper focuses on quadruped payload carrying through a passive mechanical interface, where the interaction is shaped by the passive impedance of the arm rather than by a fully actuated manipulator.
In \cite{PACC_Paper} the authors introduced a passive-arm approach for high-payload collaborative carrying with quadruped robots. The passive arm provides a lightweight compliant connection between the robot and the payload through spring and damping elements. This design preserves payload capacity and reduces actuation complexity, while allowing relative motion between the quadruped base and the carried load. The PACC controller accounts for interaction effects through force estimation and evaluates locomotion stability using the \gls{zmp} position with respect to the support polygon~\cite{ZEROMOMENTPOINTTHIRTY2004}.
However, the influence of passive-arm impedance on locomotion stability has not been isolated across different gaits and payload conditions. In particular, changing the stiffness-damping selection for the passive interface can modify the oscillatory motion of the arm-payload subsystem, and different gait patterns can excite this subsystem in different ways. This raises a practical question: how do gait pattern, payload mass, and passive-arm stiffness-damping selection affect the ZMP margin?
This paper addresses this question through a simulation-based study of gait-dependent passive payload coupling. We use the same MPC controller as in \cite{PACC_Paper} and keep it fixed across all simulations comparing damped and underdamped passive-arm configurations across multiple gait patterns and payload masses on flat ground. The goal is to evaluate how passive-arm impedance selection affects passive-joint oscillations and how these effects are reflected in the \gls{zmp} margin with respect to the support polygon.
In this work, we use the passive-arm concept introduced in \cite{PACC_Paper} as a payload-carrying interface for a single quadruped. We do not consider a collaborative-carrying scenario with a human partner or a second robot applying external interaction forces. Instead, we focus on the interaction induced by the carried payload itself, namely the inertial and gravitational effects transmitted through the passive arm during locomotion.
\subsection{Contributions}
The contributions of this paper are:
\begin{itemize}
    \item We analyze how passive-arm stiffness-damping selection affects quadruped payload carrying using the same Model Predictive Control (MPC) controller as in PACC.

    \item We quantify how passive-joint oscillations and crawl-gait ZMP-margin metrics vary with gait, payload, and passive-arm configuration.

    \item We summarize the results in gait-payload-stiffness-damping maps that identify configurations with small ZMP-margin changes and configurations where the underdamped arm produces larger losses of ZMP margin.
\end{itemize}

\section{Passive Payload Coupling Model}
 \label{sec:passivecoupling}
\subsection{Passive arm mechanism}
\label{subsec:passive_arm_mechanism}
\begin{figure}[!ht]
    \centering
    \includegraphics[width=1\linewidth]{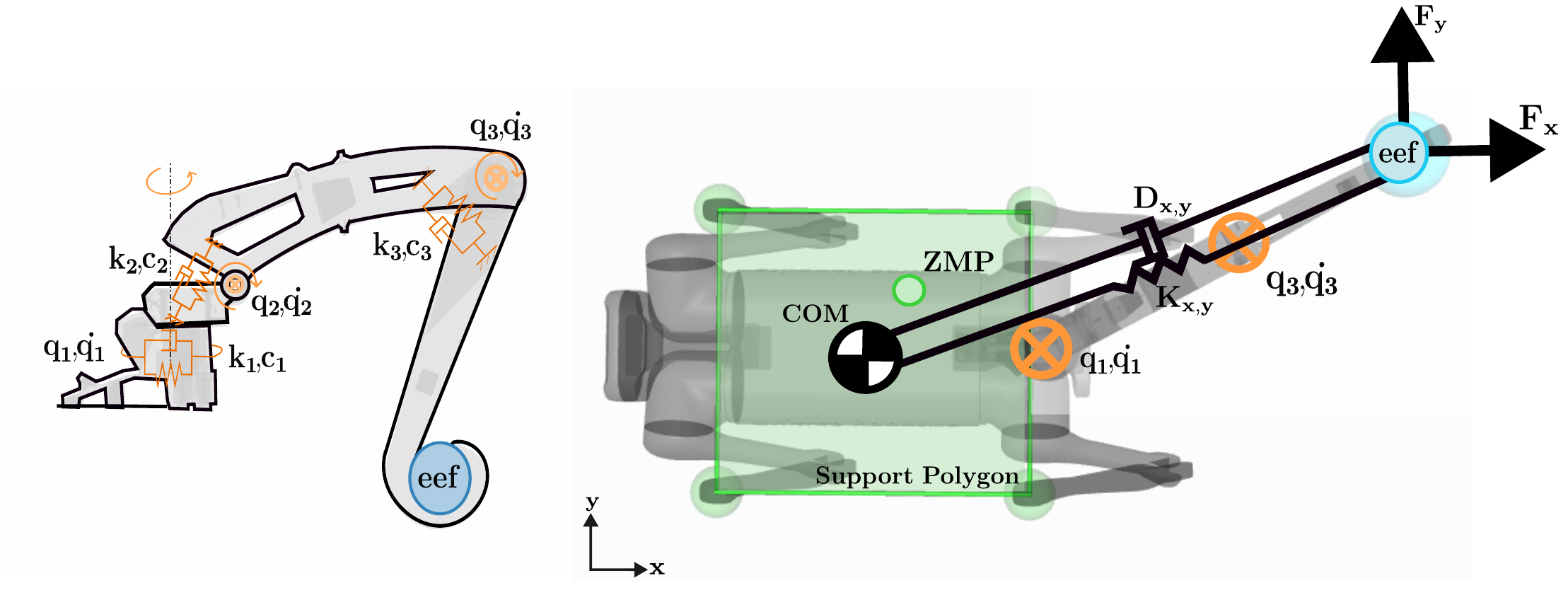}
    \caption{ 
    Left: side-view representation of the passive 3-DoF arm, showing the revolute joints and their associated stiffness and damping elements. 
    Right: top-view representation of the quadruped-passive-arm system during payload carrying. 
    The passive arm connects the robot base to the carried payload and is interpreted, as an equivalent horizontal spring-damper interface.
    }
    \label{fig:robotschematics}
\end{figure}

The passive arm considered in this work is the carrying interface introduced in PACC~\cite{PACC_Paper}. It is a lightweight 3-DoF mechanism composed of three revolute joints arranged in a yaw-pitch-pitch configuration. The joints are not actuated; their motion results from rigid-body coupling, passive elastic elements, and mechanical damping. Joint encoders measure the arm displacement, and the payload is connected to the end-effector hook.
Although PACC was originally introduced for collaborative carrying, this paper considers single-quadruped payload carrying only. Therefore, the interaction studied here is generated by the carried payload itself, through inertial and gravitational effects transmitted through the passive arm during locomotion.
The mechanism provides a compliant connection between the quadruped base and the carried load. In contrast to a rigid connection, the passive arm allows relative motion between the robot and the payload. This motion is governed by the passive joint stiffness and damping: stiffness defines the restoring behavior around the carrying configuration, while damping attenuates gait-induced oscillations of the arm-payload subsystem.
The three joints contribute differently to this coupling. Joint~1 mainly affects the horizontal orientation of the arm and lateral payload motion. Joint~2 supports the payload and regulates the nominal end-effector height. Joint~3 is the distal joint on the payload side and behaves similarly to a pendulum during locomotion. We keep the mechanism and MPC controller fixed and study how passive stiffness-damping selection affects arm oscillations and ZMP-margin behavior under different gait and payload conditions.

\subsection{Equivalent horizontal coupling model}
\label{subsec:equivalent_cartesian_interface}

Based on the mechanism described above, we approximate the passive arm-payload subsystem around the nominal carrying configuration as an equivalent two-dimensional Cartesian spring-damper interface in the horizontal plane, i.e., the plane orthogonal to gravity. Throughout this paper, the term base refers to the robot trunk. This simplified representation is not intended to model the full passive-arm dynamics, but to provide a compact way to relate gait-induced payload-base relative motion to the horizontal interaction forces transmitted through the passive interface. The equivalent system is represented in Fig.~\ref{fig:robotschematics}.

We define the horizontal payload-base displacement as
$\Delta \mathbf{x}_{xy}=\mathbf{x}_{\mathrm{load},xy}-\mathbf{x}_{\mathrm{base},xy}$,
where $\mathbf{x}_{\mathrm{load},xy}$ and $\mathbf{x}_{\mathrm{base},xy}$ are the payload and robot-base positions projected onto the horizontal plane. The corresponding horizontal interaction force is approximated as
\begin{equation}
    \mathbf{f}_{xy}
    =
    \mathbf{K}_{xy}\Delta \mathbf{x}_{xy}
    +
    \mathbf{D}_{xy}\Delta \dot{\mathbf{x}}_{xy},
    \label{eq:cartesian_spring_damper_xy}
\end{equation}
where $\mathbf{K}_{xy}$ and $\mathbf{D}_{xy}$ denote the equivalent Cartesian stiffness and damping matrices of the passive interface.
Different gait patterns and payload masses can produce different relative motions $\Delta \mathbf{x}_{xy}$ between the robot base and the carried load. The passive-arm stiffness-damping selection then determines how much of this relative motion is dissipated and how much persists as passive-joint oscillation and interaction-force variation. These effects are expected to influence the locomotion stability margin because the interaction forces, generated due to the payload inertia, are transmitted back to the robot base.
In the following subsection, we define the stability metric used to evaluate this effect. In brief, we use the \gls{zmp} margin with respect to the support polygon as an output metric of the robot locomotion stability (as in \cite{PACC_Paper}). This allows us to study whether changes in passive-arm stiffness-damping selection, gait pattern, and payload mass are reflected in the robot stability margin, without introducing a new controller or a modified \gls{zmp} formulation.
\subsection{Interaction and ZMP-margin metrics}
\label{subsec:interaction_stability_metrics}
This work uses the \gls{mpc} controller and stability metric adopted in PACC~\cite{PACC_Paper}. The controller is used without modification and is based on a reduced \gls{srbd} model and accounts for the interaction wrench generated at the passive-arm end-effector. Locomotion stability is evaluated through the \gls{zmp} position with respect to the support polygon. We do not modify this controller, introduce a new \gls{zmp} formulation, or derive a closed-form relation between passive-arm impedance and \gls{zmp} behavior.
Let $\mathcal{S}(t)$ denote the support polygon, defined as the convex hull of the stance feet at time $t$, and let $\mathbf{p}_{\mathrm{zmp}}(t)$ be the \gls{zmp} position projected on the ground plane. We define the \gls{zmp} margin $m_{\mathrm{zmp}}(t)$ as the signed distance between $\mathbf{p}_{\mathrm{zmp}}(t)$ and the boundary of $\mathcal{S}(t)$:
\begin{equation}
    m_{\mathrm{zmp}}(t)=\mathrm{sdist}\left(\mathbf{p}_{\mathrm{zmp}}(t), \partial \mathcal{S}(t)\right),
    \label{eq:zmp_margin}
\end{equation}
where $m_{\mathrm{zmp}}(t)>0$ indicates that the \gls{zmp} lies inside the support polygon, $m_{\mathrm{zmp}}(t)=0$ indicates that it lies on the boundary, and $m_{\mathrm{zmp}}(t)<0$ indicates a violation of the support polygon. In this paper, this margin is used only as an output metric of the \gls{mpc} controller.
For each simulation run, corresponding to one combination of gait pattern, payload mass, and passive-arm stiffness-damping configuration, we evaluate three groups of quantities. 
First, we log the passive-interface interaction through the horizontal end-effector force.
Second, we quantify the passive-arm response through the oscillatory motion of joints $q_1$, $q_2$, and $q_3$, computed after discarding the initial transient.
Third, we evaluate the locomotion response using the \gls{zmp} margin, including the percentage of time for which $m_{\mathrm{zmp}}$ falls below a prescribed near-threshold value and the percentage of time for which $m_{\mathrm{zmp}}<0$.

Together, these quantities evaluate the chain
$\mathrm{gait} \rightarrow \Delta \mathbf{x}_{xy} \rightarrow \mathbf{f}_{xy} \rightarrow m_{\mathrm{zmp}}$.
This chain expresses the cause-effect relation studied in this work: a gait pattern induces payload-base relative motion, the passive interface transforms this motion into arm oscillations and interaction forces, and these effects can be reflected in the ZMP margin with respect to the support polygon.

\section{Simulation Protocol}
\label{sec:Simulation Protocol}

\subsection{Platform and controller}

\label{subsec:simulation_protocol}

All simulations are performed on flat ground using the quadruped passive-arm platform described in PACC~\cite{PACC_Paper}. The platform consists of a $23\,\mathrm{kg}$ Unitree Aliengo quadruped equipped with a $1.5\,\mathrm{kg}$ passive arm mounted at the front of the robot. In the nominal standing configuration, the robot dimensions are approximately $0.8\,\mathrm{m}$ in length, $0.35\,\mathrm{m}$ in width, and $0.6\,\mathrm{m}$ in height. The simulations are performed in MuJoCo~\cite{Mujoco}.
Using the same MPC controller as in~\cite{PACC_Paper}, the commanded velocity, terrain, and nominal carrying configuration are kept fixed across the different simulations. The robot is commanded to walk forward at $0.1\,\mathrm{m/s}$ with zero yaw-rate command (heading velocity). By keeping the controller and locomotion command fixed, the analysis isolates the effects of gait pattern, payload mass, and passive-arm impedance without introducing an additional impedance-aware controller or a modified \gls{zmp} formulation.
A simulation run is defined as one execution of the controller for a specific combination of gait pattern, payload mass, passive-arm stiffness-damping configuration, and repetition index. Each run has a duration of $60\,\mathrm{s}$, and each condition is repeated three times. The objective of the simulation study is to evaluate how different locomotion patterns excite the passive arm-payload subsystem under different passive stiffness-damping configurations. In particular, we study how gait-induced relative motion between the robot base and the payload produces passive-joint oscillations and how these oscillations are reflected in the \gls{zmp} margin with respect to the support polygon.

\subsection{Gaits, payloads, and passive-arm configurations}
\label{subsec:experiment_matrix}

In this section we build an assessment matrix by varying three factors: gait pattern, payload mass, and passive-arm impedance. The four legs of the quadruped robot are denoted as left-front (LF), right-front (RF), left-hind (LH), and right-hind (RH). Four gait patterns are considered: \gls{bdc} (periodic sequence LF-RH-RF-LH), \gls{cdc} (LF-RF-LH-RH), \gls{cc} (LF-RF-RH-LH), and trot (LF/RH-RF/LH).
The three crawl gaits are selected as the main \gls{zmp}-margin comparison because their contact sequences are tuned to keep at least three feet in contact with the ground during locomotion. This provides a finite support polygon and allows the ZMP margin computation against the support polygon to be evaluated consistently. The trot gait is included as an additional dynamic reference case with a different base-motion profile. During trot, the diagonal support phase degenerates to a two-point support line, so the resulting \gls{zmp}-margin signal is not used as a direct stability metric.

The payload conditions are no-load, $1\,\mathrm{kg}$, $2\,\mathrm{kg}$, and $5\,\mathrm{kg}$.
For each gait and payload, two passive-arm configurations are tested: a damped reference configuration and an underdamped configuration. For a given payload condition, the underdamped configuration uses the same $q_2$ stiffness as the damped reference, while reducing both stiffness and damping at joints $q_1$ and $q_3$. The damping value at joint $q_2$ is also kept unchanged. The stiffness of joint $q_2$ is selected according to the payload mass to support the carried load: $k_2=5.0\,\mathrm{N\,m/rad}$ for $1\,\mathrm{kg}$, $k_2=10.0\,\mathrm{N\,m/rad}$ for $2\,\mathrm{kg}$, and $k_2=25.0\,\mathrm{N\,m/rad}$ for $5\,\mathrm{kg}$. This comparison is used to evaluate how passive-arm configurations affect payload-base oscillations and the corresponding ZMP-margin behavior with respect to the support polygon (see Sec.~\ref{sec:results}).

\begin{table}[!ht]
\centering
\caption{Passive-arm joint stiffness and damping parameters used for the damped and underdamped configurations. The value of $k_2$ is payload-dependent.}
\label{tab:arm_impedance}
\begin{tabular}{lcccccc}
\hline
\textbf{Configuration} & $k_1$ & $k_2$ & $k_3$ & $d_1$ & $d_2$ & $d_3$ \\ 
& \multicolumn{3}{c}{[$\mathrm{N\,m/rad}$]} & \multicolumn{3}{c}{[$\mathrm{N\,m\,s/rad}$]} \\
\hline
Damped & 10.0 & payload-dependent & 10.0 & 1.0 & 1.5 & 1.0 \\
Underdamped & 0.5 & payload-dependent & 0.1 & 0.05 & 1.5 & 0.05 \\
\hline
\end{tabular}
\end{table}

Each test case denotes one combination of gait pattern, payload mass, and passive-arm stiffness-damping configuration; each test is repeated three times to obtain statistical results and evaluate average and dispersion metrics over multiple runs.
The resulting matrix contains four gait patterns, four payload conditions, and two passive-arm configurations, with three repetitions for each test case.

\subsection{Data processing and computed metrics}
\label{subsec:logged_metrics}

For each simulation run, we log the robot base motion, payload motion, passive-arm joint motion, end-effector force, and \gls{zmp} margin with respect to the support polygon. The initial transient is discarded, and all metrics are computed over the remaining post-settling interval.
The passive-arm response is quantified using the oscillation RMS of joints $q_1$, $q_2$, and $q_3$, computed after removing the post-settling reference value of each joint. 

For the design-map summary, we also define a horizontal-arm oscillation index using the joints most directly associated with horizontal payload motion:
\begin{equation}
    A_{13}
    =
    \sqrt{
    \mathrm{RMS}(q_1)^2
    +
    \mathrm{RMS}(q_3)^2
    }.
    \label{eq:arm_oscillation_index}
\end{equation}
Joint $q_2$ is still reported in the joint-level oscillation comparison, but it is not included in $A_{13}$ because it mainly contributes to payload support and nominal end-effector height. The index $A_{13}$ is therefore used only as a compact summary of the passive-arm motion most related to horizontal payload-base coupling.

The locomotion response is evaluated using the \gls{zmp} margin $m_{\mathrm{zmp}}$ defined in Sec.~\ref{subsec:interaction_stability_metrics}. In particular, we compute the mean and minimum \gls{zmp} margin, the percentage of time for which $m_{\mathrm{zmp}}$ falls below the near-threshold value of $0.04\,\mathrm{m}$, and the percentage of time for which $m_{\mathrm{zmp}}<0$, corresponding to a violation of the support polygon. The value $0.04\,\mathrm{m}$ is used as a near-boundary diagnostic threshold rather than as a hard stability limit. It corresponds to about $11\%$ of the nominal robot width and $5\%$ of the nominal robot length in the standing configuration. Therefore, the metric $\%\,t(m_{\mathrm{zmp}}<0.04\,\mathrm{m})$ measures how often the robot operates with a small remaining ZMP margin, i.e., close to the support-polygon boundary. In contrast, $\%\,t(m_{\mathrm{zmp}}<0)$ measures actual support-polygon violations, where the ZMP lies outside the polygon. This distinction is important because the support polygon changes during the gait cycle: the near-threshold metric captures loss of margin before violation, while the negative-margin metric captures the most critical events.

\section{Results}
\begin{figure}[!ht]
    \centering
    \includegraphics[width=0.9\linewidth]{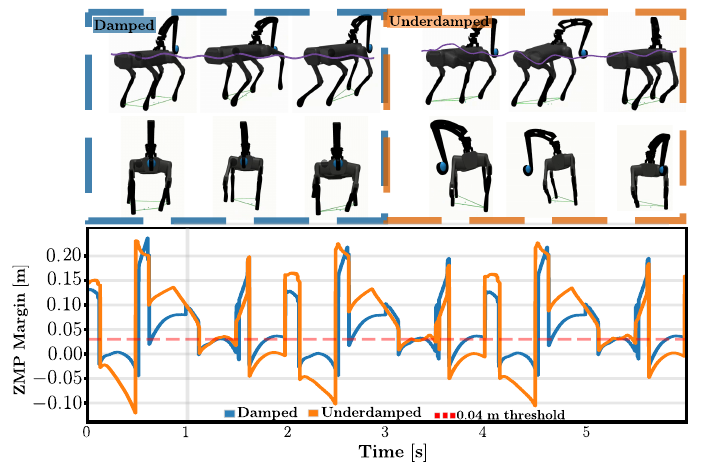}
\caption{
Representative damped and underdamped passive-arm behavior during payload-carrying circular crawl locomotion with a $5\,\mathrm{kg}$ load. The underdamped configuration produces larger arm motions and stronger ZMP-margin excursions.
}    \label{fig:arm_UD_vs_CD}
\end{figure}

Figure~\ref{fig:arm_UD_vs_CD} provides a representative example of the effect of passive-interface tuning during payload-carrying locomotion. The underdamped configuration produces larger passive-arm oscillations than the damped reference, and the corresponding ZMP-margin trace shows larger excursions.
\label{sec:results}
\subsection{ZMP-margin degradation}
\label{subsec:zmp_margin_degradation}

\begin{figure}[!ht]
    \centering
    \includegraphics[width=0.9\linewidth]{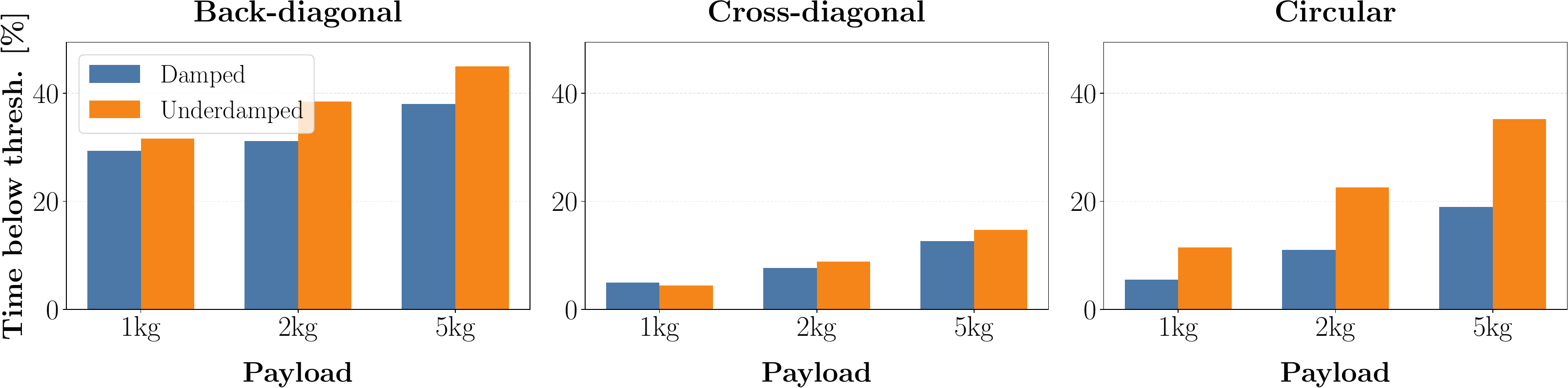}
    \caption{ZMP-margin loss for the crawl gaits. The bars report the percentage of the post-settling interval for which the ZMP margin is below $0.04\,\mathrm{m}$, indicating operation close to the support-polygon boundary. The underdamped configuration affects this percentage in a gait- and payload-dependent way, with the largest increase observed for \gls{cc} at high payload. Trot is excluded from this comparison because its diagonal support phase produces a degenerate support region.}
    \label{fig:zmp_margin_degradation}
\end{figure}

Figure~\ref{fig:zmp_margin_degradation} reports the effect of gait pattern, payload mass, and passive-arm configuration on the \gls{zmp} margin with respect to the support polygon. The bars show the percentage of time for which the \gls{zmp} margin satisfies $m_{\mathrm{zmp}}<0.04\,\mathrm{m}$, corresponding to operation close to the support polygon boundary.

The results show that the effect of passive-arm configuration is strongly dependent on the crawl footfall sequence. For \gls{bdc}, the underdamped configuration produces a moderate increase in the percentage of time with $m_{\mathrm{zmp}}<0.04\,\mathrm{m}$ as the payload increases. \gls{cdc} is the least sensitive case: even at the highest payload, the underdamped configuration produces only a small increase in the time spent below this near-threshold margin. \gls{cc} is the most sensitive case, with the largest increase in time spent near the support-polygon boundary observed at $5\,\mathrm{kg}$.

The trot gait is excluded from the \gls{zmp}-margin degradation analysis because its diagonal support phase produces a degenerate support region. In this case, the computed distance to the support boundary is dominated by the two-point support geometry and is not comparable to the finite-area support polygons obtained during crawl locomotion. Trot is therefore analyzed only through passive-arm oscillation metrics in Sec.~\ref{subsec:passive_arm_motion}. 

Actual \gls{zmp}-margin violations, defined by $m_{\mathrm{zmp}}<0$, remain close to zero for most low-payload conditions. They become relevant mainly in the underdamped high-payload crawl cases, as summarized later in the design map. These results indicate that the near-threshold metric captures the progressive loss of \gls{zmp} margin, while negative-margin events occur only in the most demanding gait-payload-stiffness-damping  combinations.

Overall, Fig.~\ref{fig:zmp_margin_degradation} shows that the underdamped configuration does not affect all crawl sequences in the same way. CDC has the smallest increase in the percentage of time spent near the support-polygon boundary, while CC has the largest increase in this metric at high payload. This indicates that the effect of passive-arm impedance depends not only on payload mass, but also on how each footfall sequence maps the payload-induced interaction into the support polygon.
The CDC result suggests that the same passive-arm excitation can have different consequences depending on the support-polygon evolution. CDC appears to keep the ZMP farther from the support-polygon boundary during critical parts of the arm-payload oscillation, while BDC and CC produce larger increases in near-threshold or negative-margin time. This supports the interpretation that the gait affects both the arm excitation and how the resulting interaction is projected onto the available ZMP margin.


\subsection{Passive-arm motion}
\label{subsec:passive_arm_motion}
\begin{figure}[!ht]
    \centering
    \includegraphics[width=0.95\linewidth]{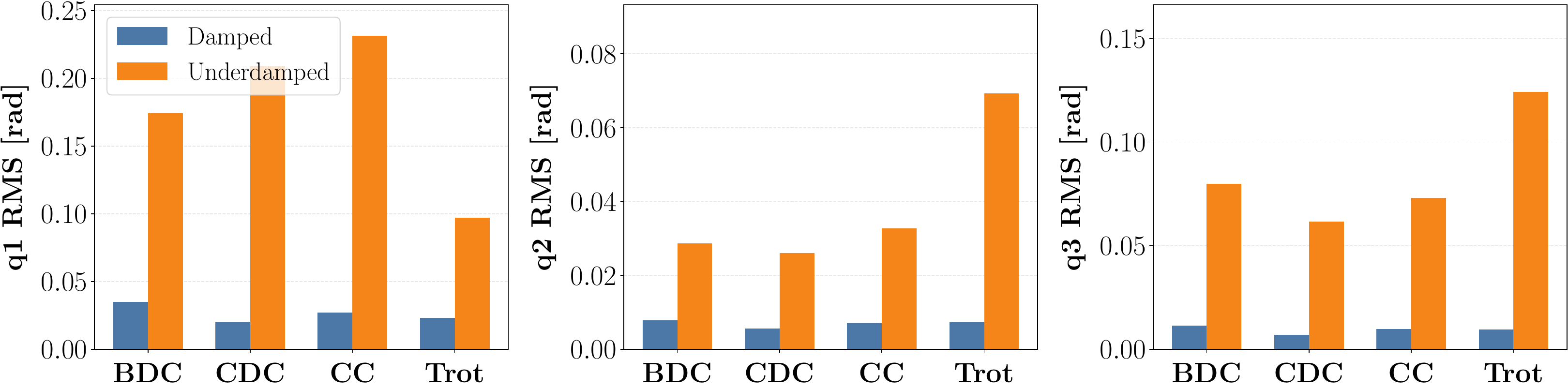}
    \caption{
    Passive-arm oscillation RMS at $2\,\mathrm{kg}$ for the tested gaits. 
    The oscillation is computed after removing the post-settling reference value of each joint. 
    The underdamped configuration increases the oscillatory motion of the passive joints, showing that reduced stiffness-damping allows gait-induced payload motion to persist in the passive arm.
    }
    \label{fig:arm_oscillation}
\end{figure}

Figure~\ref{fig:arm_oscillation} reports the passive-arm oscillation RMS for a $2\,\mathrm{kg}$ payload across the tested gaits. Unlike the ZMP-margin metric, this quantity describes the motion of the passive mechanism and is therefore also reported for trot. The oscillation RMS is computed after removing the post-settling reference value of each joint, so the plotted values represent dynamic motion around the carrying configuration.
The underdamped configuration increases passive-joint oscillations for all tested gaits. 
At $2\,\mathrm{kg}$, the combined $q_1$ and $q_3$ oscillation index increases by approximately $0.155\,\mathrm{rad}$ for \gls{bdc}, $0.196\,\mathrm{rad}$ for \gls{cdc}, $0.214\,\mathrm{rad}$ for \gls{cc}, and $0.133\,\mathrm{rad}$ for trot. \gls{cc} therefore produces the largest arm-oscillation increase among the crawl gaits, while \gls{cdc} shows a comparable arm-oscillation increase but much smaller ZMP-margin degradation. This indicates that the stability-margin effect is not determined by arm-oscillation magnitude alone, but also by how each crawl sequence maps the payload-induced interaction into the support polygon.

\subsection{Gait-payload-stiffness-damping design map}
\label{subsec:design_map}

\begin{figure}[!ht]
    \centering
    \includegraphics[width=1\linewidth]{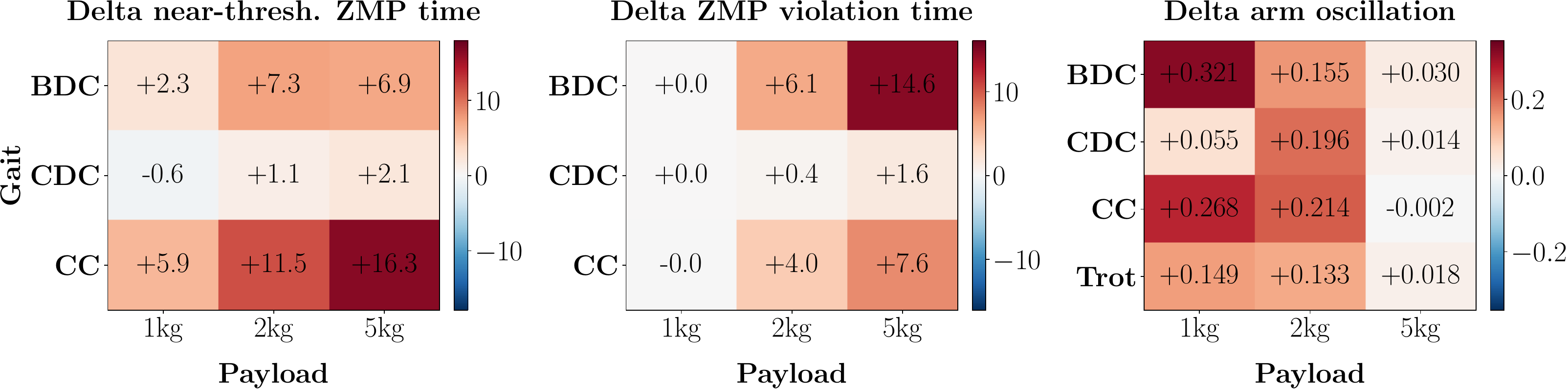}
\caption{ Gait-payload-stiffness-damping design map. Each cell reports $\Delta=\mathrm{Underdamped}-\mathrm{Damped}$ averaged over repetitions. The left and middle panels report ZMP-margin degradation for the crawl gaits, where the support polygon is finite: time with $m_{\mathrm{zmp}}<0.04\,\mathrm{m}$ and time with $m_{\mathrm{zmp}}<0$, respectively. The right panel reports the passive-arm oscillation index $A_{13}$ and includes trot as a dynamic excitation case. Positive values indicate an increase caused by the underdamped configuration. }

    \label{fig:design_map}
\end{figure}

Figure~\ref{fig:design_map} summarizes the effect of the underdamped passive-arm configuration relative to the damped reference across gait and payload conditions. Each cell reports
$\Delta=\mathrm{Underdamped}-\mathrm{Damped}$, averaged over repetitions. Positive values indicate that the underdamped configuration increases the corresponding metric.
The left panel reports the change in time for which $m_{\mathrm{zmp}}<0.04\,\mathrm{m}$. The underdamped configuration increases this near-threshold time mainly for BDC and CC, while CDC remains much less sensitive, with changes of approximately $-0.6\%$, $+1.1\%$, and $+2.1\%$ for $1\,\mathrm{kg}$, $2\,\mathrm{kg}$, and $5\,\mathrm{kg}$, respectively.
The middle panel reports the change in negative-margin time, $m_{\mathrm{zmp}}<0$. These violations remain small for most low-payload conditions and increase mainly in high-payload crawl cases, with the largest increase for BDC at $5\,\mathrm{kg}$.
The right panel reports the change in the oscillation index $A_{13}$ defined in Eq.~\eqref{eq:arm_oscillation_index}. The contrast between CDC and the other crawl gaits shows that larger passive-arm oscillation does not necessarily imply a larger loss of ZMP margin; the effect also depends on the crawl footfall sequence.
Overall, Fig.~\ref{fig:design_map} shows that passive-arm configuration, gait pattern, and payload mass should be selected jointly. CDC provides an important contrast: despite comparable passive-arm oscillation at $2\,\mathrm{kg}$, it shows smaller near-threshold and negative-margin changes, confirming that the ZMP-margin effect also depends on the crawl footfall sequence.

\section{Conclusion}
\label{sec:Conclusion}
This paper presented a simulation-based analysis of how passive-arm stiffness-damping selection affects quadruped payload carrying under different gait and payload conditions. Using the same \gls{mpc} controller as in~\cite{PACC_Paper}, we compared damped and underdamped passive-arm configurations on flat ground and evaluated their effect on passive-joint oscillations and \gls{zmp}-margin behavior.

The results show that reducing stiffness and damping increases arm oscillations and can reduce the available \gls{zmp} margin, but the magnitude of this effect depends on the gait and payload. The crawl-gait results show that sensitivity to the underdamped configuration depends on both payload mass and footfall sequence. CDC shows the smallest increase in time spent near the support-polygon boundary, CC shows the largest increase in the near-threshold \gls{zmp}-margin metric, and BDC shows the largest increase in negative-margin events at high payload. Trot is retained as a dynamic excitation case for the passive arm, but it is not used for direct \gls{zmp}-margin stability comparison because its diagonal support phase produces a two-point support region.

The gait-payload-stiffness-damping map summarizes these trends and shows that passive-interface configuration and gait selection should be considered jointly. Future work will extend the analysis to a wider range of impedance values, develop an analytical relation between gait-induced excitation, passive-arm impedance, and \gls{zmp}-margin reduction, and investigate how this relation can be used in impedance-aware predictive control. The resulting models and tuning guidelines will also be validated in real-robot experiments to assess how the simulation trends transfer to hardware.

\backmatter

\bibliography{bibliography}

\end{document}